\documentclass[journal]{IEEEtran}

\usepackage{cite}
\usepackage[cmex10]{amsmath}
\usepackage{graphicx}

\newtheorem{proposition}{Proposition}
\newtheorem{corollary}{Corollary}
\newtheorem{remark}{Remark}

\begin{document}
%
\title{3-D Velocity Regulation for Nonholonomic Source Seeking Without Position Measurement \thanks{This work was supported by National Natural Science Foundation of China (61304038, 41427806).}}
%
%
%

\author{Jinbiao~Lin,
        Shiji~Song,~\IEEEmembership{Member,~IEEE,}
        Keyou~You,~\IEEEmembership{Member,~IEEE,}
        and~Cheng~Wu
\thanks{The authors are with the Department of Automation, Tsinghua University, Beijing, 100084, China (e-mail: linjb11@mails.tsinghua.edu.cn; shijis@tsinghua.edu.cn; youky@tsinghua.edu.cn; wuc@tsinghua.edu.cn).}}

%
%

\markboth{Draft submitted to IEEE TRANSACTIONS ON CONTROL SYSTEMS TECHNOLOGY}%
{Lin \MakeLowercase{\textit{et al.}}: 3-D Velocity Regulation for Nonholonomic Source Seeking Without Position Measurement}
%


\maketitle

\begin{abstract}
We consider a three-dimensional problem of steering a nonholonomic vehicle to seek an unknown source of a spatially distributed signal field without any position measurement. In the literature, there exists an extremum seeking-based strategy under a constant forward velocity and tunable pitch and yaw velocities. Obviously, the vehicle with a constant forward velocity may exhibit certain overshoots in the seeking process and can not slow down even it approaches the source.  To resolve this undesired behavior, this paper proposes a regulation strategy for the forward velocity along with the pitch and yaw velocities. Under such a strategy, the vehicle slows down near the source and stays within a small area as if it comes to a full stop, and controllers for angular velocities become succinct. We prove the local exponential convergence via the averaging technique.  Finally, the theoretical results are illustrated with simulations.
\end{abstract}


%
\IEEEpeerreviewmaketitle

\section{Introduction}
\IEEEPARstart{R}{ecently} there has been a growing interest in the study of steering single or multiple autonomous agents to seek the source of a scalar signal field. The signal could be thermal, electromagnetic, acoustic, or the concentration of a chemical or biological agent. The strength of the signal field is usually assumed to decay away from the source, and the agent only has the capability of measuring the signal strength at its present location. However, the information about the spatial distribution of the signal field is unavailable. Such a source seeking problem is of interest in a wide range of applications, including explosive detection, drug detection, localizing the sources of hazardous chemicals leakage or pollutants, localizing hydrothermal vents, etc.

Various methods have been proposed to study source seeking problems and related issues. Porat and Nehorai \cite{Porat1996} explored the use of moving sensors for detecting and localizing vapor-emitting sources by computing the gradient of the Cram\'{e}r-Rao bound on the location  error. Ogren et al. \cite{Ogren2004} addressed the gradient climbing question of a mobile sensor network. Pang and Farrell \cite{pang2006chemical} provided a source-likelihood mapping approach based on Bayesian inference methods to estimate a
likelihood map for the location of the source of a chemical plume. Demetriou et al. \cite{Demetriou2014} proposed a coupled controls-computational fluids approach for the estimation of the signal field in a two-dimensional (2-D) domain with a Lyapunov-guided sensing aerial vehicle.

Different from \cite{Porat1996,Ogren2004,pang2006chemical,Demetriou2014} where the agents are also assumed to be capable of sensing their positions, this paper considers the three-dimensional (3-D) problem of steering a single nonholonomic vehicle to seek the source without the vehicle's position information. This consideration is motivated by vehicles operated in environments where their position information is unavailable or costly, such as urban, underground and underwater environments. The lack of position information renders the guidance of the vehicle interesting and challenging, as most of traditional searching strategies fail to work under this framework. In addition, we use only one vehicle in the mission in contrast with \cite{Porat1996,Ogren2004}. While the collaboration among multiple vehicles may improve the search efficiency due to the capacity in exploring several locations simultaneously and exchanging data in real time, a single vehicle can collect field values more freely with lower hardware overhead. As in \cite{Zhang2007,Cochran2009a,Cochran2009b,Liu2010,Liu2012,Ghods2010}, this paper focuses on seeking a source whose signal strength depends only on the distance between the vehicle and the source, .

Extremum seeking is a real-time model free gradient estimation method to optimize the steady-state output of a nonlinear system, without any explicit knowledge about the input-output map other than the existence of an extremum \cite{Tan2010}. It has been implemented successfully in many different areas \cite{Peterson2004,Li2005,Guay2005,Zhang2006,Becker2007,dower2008extremum,Ma2014,Dincmen2014}. Specially, a series of studies suggest that it is an effective approach for source seeking problems without position information \cite{Zhang2007,Cochran2009a,Cochran2009b,Liu2010,Liu2012,Ghods2010}. Employing the extremum seeking method, two distinct strategies for source seeking problems in a plane were designed. While Zhang et al. \cite{Zhang2007} considered controlling the vehicle by keeping the angular velocity constant and tuning the forward velocity, Cochran and Krstic \cite{Cochran2009b} considered controlling the vehicle by keeping the forward velocity constant and tuning the angular velocity. Liu and Krstic \cite{Liu2010,Liu2012} replaced the periodic sinusoidal excitation with the stochastic excitation and redesigned the extremum seeking method. Ghods and Krstic \cite{Ghods2010} regulated both forward velocity and angular velocity with the intent of bringing the vehicle to a stop near the source.

While most of the previous research focused on source seeking problems in a plane, there are few work on the three dimension case \cite{Cochran2009a,Matveev2014}. The 3-D model is more complicated due to more optional vehicle movements, thus the methods in the 2-D workspace can not be directly applied to the 3-D workspace. However, in many engineering applications source seeking in the 3-D workspace attracts more concern. Two control schemes in the 3-D workspace were presented by Cochran et al. \cite{Cochran2009a}. Both schemes considered a vehicle with a constant forward velocity. Whereas it is similar to the 2-D case, a constant forward velocity results in complicated asymptotic behaviors of the vehicle. As a result, the vehicle cannot settle when it approaches close to the source. Instead it exhibits certain overshoots and finally revolves around the source with a relatively large forward velocity. Note that a small constant forward velocity may improve the asymptotic performance, but the convergence rate decreases. Matveev et al. \cite{Matveev2014} proposed a hybrid controller without estimation of the field gradient, which is justified to be of global convergence to the source of a generic field. However, they employed a vehicle with a constant forward velocity as well, which again results in the undesired performance similar to \cite{Cochran2009a}.

To overcome the shortcoming induced by a constant forward velocity in the 2-D case, an approach that regulates both forward velocity and angular velocity was proposed by Ghods and Krstic \cite{Ghods2010}. In this paper the approach is extended to 3-D workspace, aiming at bringing the vehicle to a stop around the source. While only one angular velocity is considered in the 2-D case, we must address two angular velocities (pitch and yaw velocities) simultaneously in the 3-D case. The two angular velocities and the forward velocity interwine with each other, which makes the design and analysis of the control scheme challenging. A preliminary version of this work has been reported in WCICA 2014 \cite{Lin2014}. However, we establish a rigorous proof of the stability and provide a rule on parameter selection for the controllers in this paper.

We tune both angular velocities and forward velocity simultaneously in contrast with \cite{Cochran2009a,Matveev2014}. Under a tunable forward velocity we can slow down the vehicle around the source to get closer to the source without decreasing the convergence rate. The forward velocity is controlled to be small near the source, which eliminates the undesired overshoots. Moreover, the controllers for angular velocities can be simplified by removing extra items for eliminating the overshoots caused by a constant large forward velocity. We obtain two phenomena for the static source that produces a signal field with spherical level sets. In one phenomenon the vehicle eventually stays within a small area around the source as if it ``stops" as expected. In the other it revolves in an annular attractor around the source, which is similar to \cite{Cochran2009a} but with simpler controllers.

The rest of the paper is organized as follows. In Section 2 we describe the 3-D nonholonomic source seeking problem and present our extremum seeking scheme. In Section 3 we derive an averaged system and prove local exponential convergence for the static source that produces a signal field with spherical level sets. In Section 4 we provide a rule on the parameter selection for the control scheme. In Section 5 we include simulation results to illustrate the behaviors of the vehicle under different scenarios.

\section{Problem Description and Control Scheme}
In this section we firstly describe the 3-D nonholonomic vehicle model for the task of the source seeking mission. Then we present our extremum seeking scheme and provide an intuitive interpretation of the control scheme.
\subsection{Problem Description}
\begin{figure}[!t]
  \centering
  \includegraphics[width=0.8\hsize]{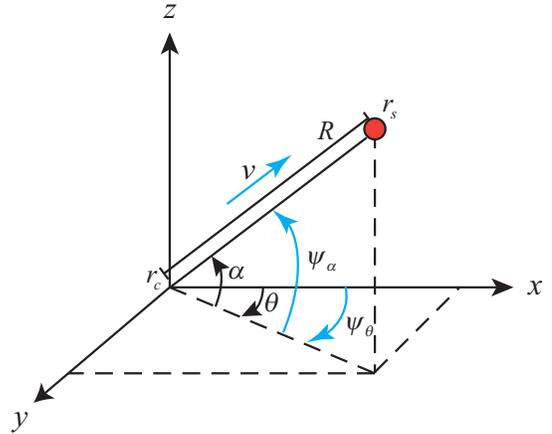}
  \caption{Geometric interpretation of vehicle model.}
  \label{fig1}
\end{figure}

As in \cite{Cochran2009a}, we consider an autonomous vehicle under nonholonomic constraints,  see Fig.~\ref{fig1} for an illustration. The vehicle is equipped with actuators which are used to impart forward, pitch and yaw velocities. The diagram in Fig.~\ref{fig1} depicts the position, heading, forward, pitch and yaw velocities for the vehicle center and sensor. The sensor $r_s$ is mounted at a distance $R$ away from the vehicle center $r_c$. The azimuthal angle $\alpha$ defines the pitch, whose velocity is governed by $\psi_\alpha$. The polar angle $\theta$ defines the yaw, whose velocity is governed by $\psi_\theta$. The forward velocity, or surge velocity, is defined by $v$. Note that the sensor is mounted at the tip of the vehicle, thus the roll velocity and angle do not affect the measurement and movement of the vehicle and can be neglected.

The kinematic equations of motion for the vehicle center are
\begin{eqnarray}
{{\dot r}_c} &=& v\left[ {\begin{array}{*{20}{c}}
{\cos (\alpha )\cos (\theta )}\\
{\cos (\alpha )\sin (\theta )}\\
{\sin (\alpha )}
\end{array}} \right], \label{eq:motion1}  \\
\dot \alpha  &=& {\psi _\alpha }, \label{eq:motion2} \\
\dot \theta  &=& {\psi _\theta }, \label{eq:motion3}
\end{eqnarray}
where $r_c = (x_c,y_c,z_c)$. The sensor is located at
\begin{equation}
{r_s} = {r_c} + R\left[ {\begin{array}{*{20}{c}}
{\cos (\alpha )\cos (\theta )}\\
{\cos (\alpha )\sin (\theta )}\\
{\sin (\alpha )}
\end{array}} \right].  \label{eq:motion4}
\end{equation}

The task of vehicle is to seek a source that emits a spatially distributed signal. The signal strength at the location $r$ is denoted by  $J = f\left( r \right)$, which has an isolated local maximum $f^* = f(r^*)$ where $r^*$ denotes the source location, or the local maximizer of $f$. Here the isolated maximum means that there is only one source in the searching area. The signal strength $J$ decays away from the source $r^*$, but other information about $J$, such as the shape of $f$ and the position of $r^*$, is unknown. Furthermore, the information of the vehicle's position is unavailable, which makes traditional gradient searching strategy fail to work. Our objective of this work is to design a control scheme to steer the nonholonomic vehicle to the source without any vehicle's position information.

\subsection{Control Scheme}
To achieve the goal of seeking the source, we employ an extremum seeking method to tune the pitch and yaw velocities ($\psi _\alpha$ and $\psi _\theta$) directly and the forward velocity ($v$) indirectly, which is in contrast with \cite{Cochran2009a} as they assume the forward velocity to be constant. It is worth mentioning that the 2-D case has been studied in \cite{Ghods2010}. However, the 3-D case is more intricate and challenging and remains outstanding. Firstly, the control scheme in \cite{Ghods2010} for the 2-D case is no longer applicable. We consider more angular variables while the utilizable information is still constrained to the signal strength. In fact two extra states are added to analyze the dynamics of the system for the extra dimension. Secondly, the two angular velocities and the forward velocity interact with each other by affecting the motion and movement of the vehicle. The coupling of the three velocities obviously complicates the stability analysis of the vehicle dynamics and involves new challenges. Thus, we need to redesign the control scheme.

\begin{figure}[!t]
  \centering
  \includegraphics[width=0.9\hsize]{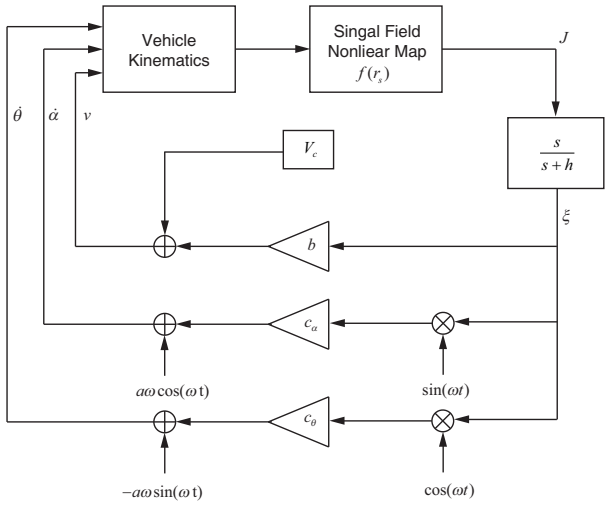}
  \caption{Block diagram of source seeking via tuning of forward, pitch and yaw velocities of the vehicle.}
  \label{fig2}
\end{figure}

Our control scheme is depicted by the block diagram in Fig.~\ref{fig2}. The control laws are given by
\begin{eqnarray}
v &=& {V_c} + b\xi, \label{eq:law1}\\
{\psi _\alpha }  &=& a\omega \cos (\omega t) + {c_\alpha }\xi \sin (\omega t), \label{eq:law2}\\
{\psi _\theta }  &=&  - a\omega \sin (\omega t) + {c_\theta }\xi \cos (\omega t), \label{eq:law3}\\
\xi  &=& \frac{s}{{s + h}}[J], \label{eq:law4}
\end{eqnarray}
where $\omega$ is the probing frequency, the parameters $a$, $c_\alpha$, $c_\theta$, $b$ and $V_c$ are positive and will affect the performance of the approach, $J$ is the sensor reading, and $\xi$ is the output of the washout filter $\frac{s}{{s + h}}[J]$. In the sequel, we shall provide a rule for designing the above parameters.

In our control scheme in Fig.~\ref{fig2}, the pitch velocity $\psi_\alpha$ and the yaw velocity $\psi_\theta$ are tuned according to the idea of the basic extremum seeking tuning law \cite{Ariyur2003}. The perturbation terms, $a\omega \cos (\omega t)$ and $- a\omega \sin (\omega t)$, are added to persistently excite the system while the corresponding demodulation terms, $\xi \sin (\omega t)$ and $\xi \cos (\omega t)$, are used to estimate the gradient of the nonlinear map $f$. The washout filter $\frac{s}{s+h}$ is used to eliminate the ``DC component" of the sensor reading $J$. The forward velocity $v$ is designed to be positively correlated to the output  $\xi$ of the washout filter, since $\xi$ describes the variation of the sensor reading $J$ in some sense. As a result, the vehicle would speed up when heading towards the source, and slow down when deviating from the source. A detailed stability analysis of this scheme for the static source will be shown in next section.

\begin{figure}[!t]
  \centering
  \includegraphics[width=\hsize]{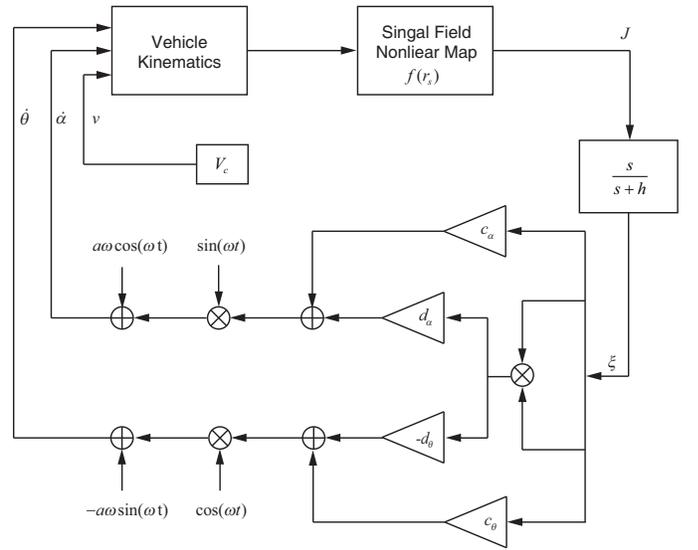}
  \caption{Block diagram of source seeking in \cite{Cochran2009a}.}
  \label{fig3}
\end{figure}

It is worthy mentioning that our control scheme is quite different from the one in \cite{Cochran2009a}, which is given in Fig.~\ref{fig3}. In their work a vehicle with a constant forward velocity is considered. Employing the basic extremum seeking method, the vehicle cannot settle even if it has reached the source due to the use of a constant forward velocity. In addition, it easily overshoots the source and turns around. The process may repeat for a while before the vehicle finally revolves around the source with a constant velocity. In \cite{Cochran2009a} a so-called ``d-item" is added to tune the angular velocities, aiming at improving the performance. Nevertheless, a constant forward velocity leads to complicated asymptotic behaviors of the vehicle.

Instead of using a constant forward velocity, we tune the forward velocity along with the pitch and yaw velocities. This intuitively is a better way to control the vehicle, as we are able to smartly adjust the vehicle to speed up, slow down, or stop depending on different circumstances. Besides, we can conveniently maneuvre the angular velocities under a tunable forward velocity, which is evident from Fig.~\ref{fig2}. While the control scheme in \cite{Cochran2009a} employed relatively complex controllers as shown in Fig.~\ref{fig3}, we can simply employ the basic extremum seeking controllers. This succinct feature substantially simplifies the parameter design of the control scheme.

\section{Stability of the Closed-loop System}

It is obvious that the dynamics of the closed-loop system is complicated due to the nonlinearities of the vehicle model and the signal map and the time varying forcing applied by extremum seeking. Additionally, more coupled state variables are involved to analyze the dynamics than in the 2-D case. In this section we focus on a static source that produces a signal field with spherical level sets and employ an averaging method \cite{Khalil2002} to analyze the stability of the system.

In this work, the distribution of the signal field is assumed to depend only on the distance from the sensor to the source. Since we are concerned with the local convergence, it is sensible to apply the Taylor series expansion around the source and omit higher order terms, which yields a quadratic map. Thus, the quadratic map is an approximation of the distribution of the signal field in a local sense, and takes the following form
\begin{equation}
J = f(r_s) = {f^*} - {q_r}{\left| {{r_s} - {r^*}} \right|^2},
\end{equation}
where $r^*$ is the unknown maximizer and denotes the location of source, $f^* = f(r^*)$ is the unknown maximum and $q_r$ is an unknown positive constant.

To analyze the stability of the closed-loop system, we define an output error variable
\begin{equation}
e = \frac{h}{{s + h}}[J] - {f^*},
\end{equation}
which allows us to express the output  of the washout filter as
$$\xi  = \frac{s}{{s + h}}[J] = J - \frac{h}{{s + h}}[J] = J - {f^*} - e.$$

Moreover, we can easily obtain that $\dot e = h\xi$. By inserting the control law (\ref{eq:law1})-(\ref{eq:law4}) into the system (\ref{eq:motion1})-(\ref{eq:motion4}), the closed-loop system is written as
\begin{eqnarray}
{{\dot r}_c} &=& ({V_c} + b\xi )\left[ {\begin{array}{*{20}{c}}
{\cos (\alpha )\cos (\theta )}\\
{\cos (\alpha )\sin (\theta )}\\
{\sin (\alpha )}
\end{array}} \right], \label{eq:sys1}\\
\dot \alpha  &=& a\omega \cos (\omega t) + {c_\alpha }\xi \sin (\omega t),\\
\dot \theta  &=&  - a\omega \sin (\omega t) + {c_\theta }\xi \cos (\omega t),\\
\dot e &=& h\xi, \\
\xi  &=&  - {q_r}{\left| {{r_s} - {r^*}} \right|^2} - e,\\
{r_s} &=& {r_c} + R\left[ {\begin{array}{*{20}{c}}
{\cos (\alpha )\cos (\theta )}\\
{\cos (\alpha )\sin (\theta )}\\
{\sin (\alpha )}
\end{array}} \right]. \label{eq:sys6}
\end{eqnarray}

The closed-loop system (\ref{eq:sys1})-(\ref{eq:sys6}) involves six variables, and it is difficult to analyze the system directly. We re-express the system by variable transformation and time scale change for further discussion firstly. Noting the signal strength $J$ depends only on the distance between the vehicle and the source, we redefine the position of the vehicle center in its spherical coordinates form. Finally we successfully reduce the system order by defining an error variable. To this end, define shifted variables by
\begin{eqnarray}
{{\hat r}_c} &=& {r_c} - {r^*}, \notag\\
\hat \alpha  &=& \alpha  - a\sin (\omega t), \notag\\
\hat \theta  &=& \theta  - a\cos (\omega t), \notag\\
\hat e &=& e + {q_r}{R^2}, \notag
\end{eqnarray}
and introduce a time scale change
\begin{equation}
\tau  = \omega t. \notag
\end{equation}
The dynamics of the shifted system is given by
\begin{align}
\frac{{d{{\hat r}_c}}}{{d\tau }} &= \frac{{{V_c} + b\xi }}{\omega }\left[ {\begin{array}{*{20}{c}}
{\cos (\hat \alpha  + a\sin (\tau ))\cos (\hat \theta  + a\cos (\tau ))}\\
{\cos (\hat \alpha  + a\sin (\tau ))\sin (\hat \theta  + a\cos (\tau ))}\\
{\sin (\hat \alpha  + a\sin (\tau ))}
\end{array}} \right], \notag\\
\frac{{d\hat \alpha }}{{d\tau }} &= \frac{{{c_\alpha }\xi }}{\omega }\sin (\tau ), \notag\\
\frac{{d\hat \theta }}{{d\tau }} &= \frac{{{c_\theta }\xi }}{\omega }\cos (\tau ), \notag\\
\frac{{d\hat e}}{{d\tau }} &= \frac{{h\xi }}{\omega }. \notag
\end{align}

For further analysis we now redefine ${\hat r}_c$ by its spherical coordinates
\begin{eqnarray}
{{\tilde r}_c} &=& \left| {{{\hat r}_c}} \right| = \sqrt {\hat x_c^2 + \hat y_c^2 + \hat z_c^2}, \notag\\
 - {{\hat r}_c} &=& {{\tilde r}_c}\left[ {\begin{array}{*{20}{c}}
{\cos ({\alpha ^*})\cos ({\theta ^*})}\\
{\cos ({\alpha ^*})\sin ({\theta ^*})}\\
{\sin ({\alpha ^*})}
\end{array}} \right], \notag\\
\tan ({\alpha ^*}) &=&  - \frac{{{{\hat z}_c}}}{{\sqrt {\hat x_c^2 + \hat y_c^2} }}, \notag\\
\tan ({\theta ^*}) &=& \frac{{{{\hat y}_c}}}{{{{\hat x}_c}}}, \notag
\end{eqnarray}
where ${\tilde r}_c$ is the distance between the vehicle center and the source, ${\alpha ^*}$ and ${\theta ^*}$ represent the azimuthal angle and polar angle towards the source when the vehicle is at ${\hat r}_c$ respectively.
By using these new definitions, the expression of $\xi$ is
\begin{align}
\xi  &=  - {q_r}(\tilde r_c^2 + {R^2} - 2{{\tilde r}_c}R{\xi _c}) - e, \notag\\
{\xi _c} &= \cos (\hat \alpha  + a\sin (\tau ))\cos ({\alpha ^*})\cos (\hat \theta- {\theta ^*}  + a\cos (\tau )) \notag \\
& \quad + \sin (\hat \alpha  + a\sin (\tau ))\sin ({\alpha ^*}), \notag
\end{align}
and the resulting dynamics is
\begin{align}
\frac{{d{{\tilde r}_c}}}{{d\tau }} &= \frac{{\frac{{d{{\hat x}_c}}}{{d\tau }}{{\hat x}_c} + \frac{{d{{\hat y}_c}}}{{d\tau }}{{\hat y}_c} + \frac{{d{{\hat z}_c}}}{{d\tau }}{{\hat z}_c}}}{{{{\tilde r}_c}}} \notag\\
&= - \frac{{({V_c} + b\xi ){\xi _c}}}{\omega }, \notag\\
\frac{{d{\alpha ^*}}}{{d\tau }} &= \frac{{{{\hat z}_c}\left( {\frac{d}{{d\tau }}\sqrt {\hat x_c^2 + \hat y_c^2} } \right) - \frac{{d{{\hat z}_c}}}{{d\tau }}\sqrt {\hat x_c^2 + \hat y_c^2} }}{{\tilde r_c^2}} \notag\\
&= - \frac{{{V_c} + b\xi }}{{\omega {{\tilde r}_c}}}\Big[ {\sin (\hat \alpha  + a\sin (\tau ))\cos ({\alpha ^*}) - } \notag\\
& \quad  { \cos (\hat \alpha  + a\sin (\tau ))\sin ({\alpha ^*})\cos (\hat \theta  - {\theta ^*} + a\cos (\tau ))} \Big], \notag\\
\frac{{d{\theta ^*}}}{{d\tau }} &= \frac{{\frac{{d{{\hat y}_c}}}{{d\tau }}{{\hat x}_c} - {{\hat y}_c}\frac{{d{{\hat x}_c}}}{{d\tau }}}}{{\hat y_c^2 + \hat x_c^2}} \notag\\
&= - \frac{{{V_c} + b\xi }}{{\omega {{\tilde r}_c}\cos ({\alpha ^*})}}\times \notag\\
& \quad \left[ {\cos (\hat \alpha  + a\sin (\tau ))\sin (\hat \theta  - {\theta ^*} + a\cos (\tau ))} \right]. \notag
\end{align}

The system order can be reduced from six to five by defining an error variable $\tilde \theta  = \hat \theta  - {\theta ^*}$, resulting in the following error system
\begin{align}
\frac{{d{{\tilde r}_c}}}{{d\tau }} &=  - \frac{{({V_c} + b\xi ){\xi _c}}}{\omega }, \label{eq:es1}\\
\frac{{d{\alpha ^*}}}{{d\tau }} &=  - \frac{{{V_c} + b\xi }}{{\omega {{\tilde r}_c}}}\Big[ {\sin (\hat \alpha  + a\sin (\tau ))\cos ({\alpha ^*})} \notag \\
& \quad {- \cos (\hat \alpha  + a\sin (\tau ))\sin ({\alpha ^*})\cos (\tilde \theta  + a\cos (\tau ))} \Big], \\
\frac{{d\hat \alpha }}{{d\tau }} &= \frac{{{c_\alpha }\xi }}{\omega }\sin (\tau ), \\
\frac{{d\tilde \theta }}{{d\tau }} &= \frac{{{c_\theta }\xi }}{\omega }\cos (\tau ) + \frac{{{V_c} + b\xi }}{{\omega {{\tilde r}_c}\cos ({\alpha ^*})}} \times \notag \\
& \quad \left[ {\cos (\hat \alpha  + a\sin (\tau ))\sin (\tilde \theta  + a\cos (\tau ))} \right],
\end{align}
\begin{align}
\frac{{d\hat e}}{{d\tau }} &= \frac{h}{\omega }\xi, \label{eq:es5}
\end{align}
where
\begin{align}
\xi  &=  - {q_r}\tilde r_c^2 + 2{q_r}R{{\tilde r}_c}{\xi _c} - \hat e, \notag\\
{\xi _c} &= \cos (\hat \alpha  + a\sin (\tau ))\cos ({\alpha ^*})\cos (\tilde \theta  + a\cos (\tau )) \notag \\
& \quad + \sin (\hat \alpha  + a\sin (\tau ))\sin ({\alpha ^*}). \notag
\end{align}

Note that the system equations are periodic in $2\pi$. According to the averaging method in \cite{Khalil2002}, the solution of the error system can be well  approximated by the solution of the corresponding ``average system". Averaging the error system in a period $2\pi$, we obtain an average error system
\begin{align}
\frac{{d\tilde r_c^{{\rm{ave}}}}}{{d\tau }} &= \frac{{(b{q_r}{{(\tilde r_c^{{\rm{ave}}})}^2} + b{{\hat e}^{{\rm{ave}}}} - {V_c})}}{\omega }\xi _c^{{\rm{ave}}} \notag\\
& \quad - \frac{{2b{q_r}R\tilde r_c^{{\rm{ave}}}}}{\omega }\xi _c^{2{\rm{ave}}}, \label{eq:ave1} \\
\frac{{d{\alpha ^*}^{{\rm{ave}}}}}{{d\tau }} &= \frac{{(b{q_r}{{(\tilde r_c^{{\rm{ave}}})}^2} + b{{\hat e}^{{\rm{ave}}}} - {V_c})}}{{\omega \tilde r_c^{{\rm{ave}}}}}{\xi ^{\genfrac{}{}{0pt}{}{\scriptstyle\alpha\hfill}{\scriptstyle{\rm{ave}}}}}
- \frac{{2b{q_r}R}}{\omega }\xi _c^{\genfrac{}{}{0pt}{}{\scriptstyle\alpha\hfill}{\scriptstyle{\rm{ave}}}},\\
\frac{{d{{\hat \alpha }^{{\rm{ave}}}}}}{{d\tau }} &= \frac{{2{c_\alpha }{q_r}R\tilde r_c^{{\rm{ave}}}}}{\omega }\xi _c^{\genfrac{}{}{0pt}{}{\scriptstyle\sin}{\scriptstyle{\rm{ave}}}},\\
\frac{{d{{\tilde \theta }^{{\rm{ave}}}}}}{{d\tau }} &= \frac{{2{c_\theta }{q_r}R\tilde r_c^{{\rm{ave}}}}}{\omega }\xi _c^{\genfrac{}{}{0pt}{}{\scriptstyle\cos}{\scriptstyle{\rm{ave}}}} + \frac{{2b{q_r}R}}{{\omega \cos ({\alpha ^{*{\rm{ave}}}})}}\xi _c^{\genfrac{}{}{0pt}{}{\scriptstyle\cos\sin}{\scriptstyle{\rm{ave}\hfill}}} \notag \\
& \quad + \frac{{({V_c} - b{q_r}{{(\tilde r_c^{{\rm{ave}}})}^2} - b{{\hat e}^{{\rm{ave}}}})}}{{\omega \tilde r_c^{{\rm{ave}}}\cos ({\alpha ^{*{\rm{ave}}}})}}{\xi ^{\genfrac{}{}{0pt}{}{\scriptstyle\cos\sin}{\scriptstyle{\rm{ave}\hfill}}}}, \\
\frac{{d{{\hat e}^{{\rm{ave}}}}}}{{d\tau }} &=  - \frac{{h{q_r}}}{\omega }{(\tilde r_c^{{\rm{ave}}})^2} - \frac{h}{\omega }{{\hat e}^{{\rm{ave}}}} + \frac{{2h{q_r}R}}{\omega }\tilde r_c^{{\rm{ave}}}\xi _c^{{\rm{ave}}}, \label{eq:ave5}
\end{align}
where ${J_0}( \cdot )$ and ${J_1}( \cdot )$ are Bessel functions of the first kind and
\begin{align}
\xi _c^{{\rm{ave}}} &= {J_0}(\sqrt 2 a)\cos ({\alpha ^*}^{{\rm{ave}}})\cos ({{\hat \alpha }^{{\rm{ave}}}})\cos ({{\tilde \theta }^{{\rm{ave}}}}) \notag \\
&\quad + {J_0}(a)\sin ({\alpha ^{*{\rm{ave}}}})\sin ({{\hat \alpha }^{{\rm{ave}}}}), \notag\\
\xi _c^{{2\rm{ave}}} &= \frac{{{{\cos }^2}({\alpha ^{*{\rm{ave}}}})}}{4} \bigg\{ {{J_0}(2\sqrt 2 a)\cos (2{{\hat \alpha }^{{\rm{ave}}}})\cos (2{{\tilde \theta }^{{\rm{ave}}}}) } \notag\\
&\quad {+ {J_0}(2a)\left[ {\cos (2{{\hat \alpha }^{{\rm{ave}}}}) + \cos (2{{\tilde \theta }^{{\rm{ave}}}})} \right] + 1} \bigg\}\notag\\
&\quad + \frac{{{{\sin }^2}({\alpha ^{*{\rm{ave}}}})}}{2}\left[ {1 - {J_0}(2a)\cos (2{{\hat \alpha }^{{\rm{ave}}}})} \right]\notag \\
&\quad + \frac{{{J_0}(\sqrt 5 a)}}{2}\sin (2{\alpha ^{*{\rm{ave}}}})\sin (2{{\hat \alpha }^{{\rm{ave}}}})\cos ({{\tilde \theta }^{{\rm{ave}}}}), \notag
\end{align}
\begin{align}
{\xi ^{\genfrac{}{}{0pt}{}{\scriptstyle\alpha\hfill}{\scriptstyle{\rm{ave}}}}} &= {J_0}(a)\cos ({\alpha ^{*{\rm{ave}}}})\sin ({{\hat \alpha }^{{\rm{ave}}}})  \notag \\
&\quad - {J_0}(\sqrt 2 a)\sin ({\alpha ^{*{\rm{ave}}}})\cos ({{\hat \alpha }^{{\rm{ave}}}})\cos ({{\tilde \theta }^{{\rm{ave}}}}), \notag\\
\xi _c^{\genfrac{}{}{0pt}{}{\scriptstyle\alpha\hfill}{\scriptstyle{\rm{ave}}}} &= \frac{{{J_0}(\sqrt 5 a)}}{2}\cos (2{\alpha ^{*{\rm{ave}}}})\sin (2{{\hat \alpha }^{{\rm{ave}}}})\cos ({{\tilde \theta }^{{\rm{ave}}}}) \notag \\
&\quad - \frac{{\sin (2{\alpha ^{*{\rm{ave}}}})}}{8} \Big[ {{J_0}(2\sqrt 2 a)\cos (2{{\hat \alpha }^{{\rm{ave}}}})\cos (2{{\tilde \theta }^{{\rm{ave}}}}) } \notag\\
&\quad { + 3{J_0}(2a)\cos (2{{\hat \alpha }^{{\rm{ave}}}}) + {J_0}(2a)\cos (2{{\tilde \theta }^{{\rm{ave}}}}) - 1} \Big], \notag\\
\xi _c^{\genfrac{}{}{0pt}{}{\scriptstyle\sin}{\scriptstyle{\rm{ave}}}} &=  - \frac{{{J_1}(\sqrt 2 a)}}{{\sqrt 2 }}\cos ({\alpha ^{*{\rm{ave}}}})\sin ({{\hat \alpha }^{{\rm{ave}}}})\cos ({{\tilde \theta }^{{\rm{ave}}}}) \notag\\
&\quad + {J_1}(a)\sin ({\alpha ^*}^{{\rm{ave}}})\cos ({{\hat \alpha }^{{\rm{ave}}}}), \notag\\
\xi _c^{\genfrac{}{}{0pt}{}{\scriptstyle\cos}{\scriptstyle{\rm{ave}}}} &=  - \frac{{{J_1}(\sqrt 2 a)}}{{\sqrt 2 }}\cos ({\alpha ^{*{\rm{ave}}}})\cos ({{\hat \alpha }^{{\rm{ave}}}})\sin ({{\tilde \theta }^{{\rm{ave}}}}), \notag\\
\xi _c^{\genfrac{}{}{0pt}{}{\scriptstyle\cos\sin}{\scriptstyle{\rm{ave}\hfill}}} &= \frac{{{J_0}(2\sqrt 2 a)}}{4}\cos ({\alpha ^{*{\rm{ave}}}})\cos (2{{\hat \alpha }^{{\rm{ave}}}})\sin (2{{\tilde \theta }^{{\rm{ave}}}})\notag\\
&\quad + \frac{{{J_0}(2a)}}{4}\cos ({\alpha ^{*{\rm{ave}}}})\sin (2{{\tilde \theta }^{{\rm{ave}}}})\notag\\
&\quad + \frac{{{J_0}(\sqrt 5 a)}}{2}\sin ({\alpha ^*}^{{\rm{ave}}})\sin (2{{\hat \alpha }^{{\rm{ave}}}})\sin ({{\tilde \theta }^{{\rm{ave}}}}), \notag\\
{\xi ^{\genfrac{}{}{0pt}{}{\scriptstyle\cos\sin}{\scriptstyle{\rm{ave}\hfill}}}} &= {J_0}(\sqrt 2 a)\cos ({{\hat \alpha }^{{\rm{ave}}}})\sin ({{\tilde \theta }^{{\rm{ave}}}}). \notag
\end{align}

By setting the right hand side of (\ref{eq:ave1})-(\ref{eq:ave5}) to be zero, we obtain the following four equilibria with details being provided in Appendix \ref{appendix_equlibira}\footnote{According to the  physical interpretation of the system, here we have implicitly restricted ${\alpha ^{*{\rm{ave}}}}\in[-\pi/2,\pi/2], {{\hat \alpha }^{{\rm{ave}}}} \in[-\pi/2,\pi/2], {{\tilde \theta }^{{\rm{ave}}}} \in (-\pi,\pi]$ to exclude repetitive equilibria.}.
\begin{align}
&\left[ {\tilde r_c^{{\rm{ave}}{^{{\rm{eq1}}}}},{\alpha ^{*{\rm{ave}}{^{{\rm{eq1}}}}}},{{\hat \alpha }^{{\rm{ave}}{^{{\rm{eq1}}}}}},{{\tilde \theta }^{{\rm{ave}}{^{{\rm{eq1}}}}}},\;{{\hat e}^{{\rm{ave}}{^{{\rm{eq1}}}}}}} \right] \notag\\
&= \left[ {{\gamma _1},0,0,0,{e_{1}}} \right], \label{eq:eq1}\\
&\left[ {\tilde r_c^{{\rm{ave}}{^{{\rm{eq2}}}}},{\alpha ^{*{\rm{ave}}{^{{\rm{eq2}}}}}},{{\hat \alpha }^{{\rm{ave}}{^{{\rm{eq2}}}}}},{{\tilde \theta }^{{\rm{ave}}{^{{\rm{eq2}}}}}},\;{{\hat e}^{{\rm{ave}}{^{{\rm{eq2}}}}}}} \right] \notag\\
&= \left[ -{{\gamma _1},0,0,\pi,{e_{1}}} \right], \label{eq:eq2}\\
&\left[ {\tilde r_c^{{\rm{ave}}{^{{\rm{eq3}}}}},{\alpha ^{*{\rm{ave}}{^{{\rm{eq3}}}}}},{{\hat \alpha }^{{\rm{ave}}{^{{\rm{eq3}}}}}},{{\tilde \theta }^{{\rm{ave}}{^{{\rm{eq3}}}}}},\;{{\hat e}^{{\rm{ave}}{^{{\rm{eq3}}}}}}} \right] \notag\\
&= \left[ {{\rho _2}\sqrt {2{\gamma _3}} ,0,0,{\mu _0},{e_{2}}} \right], \label{eq:eq3}\\
&\left[ {\tilde r_c^{{\rm{ave}}{^{{\rm{eq4}}}}},{\alpha ^{*{\rm{ave}}{^{{\rm{eq4}}}}}},{{\hat \alpha }^{{\rm{ave}}{^{{\rm{eq4}}}}}},{{\tilde \theta }^{{\rm{ave}}{^{{\rm{eq4}}}}}},\;{{\hat e}^{{\rm{ave}}{^{{\rm{eq4}}}}}}} \right] \notag\\
&= \left[ {{\rho _2}\sqrt {2{\gamma _3}} ,0,0, - {\mu _0},{e_{2}}} \right], \label{eq:eq4}
\end{align}
where
\begin{align}
{\gamma _1} &= \frac{{{V_c}{J_0}(\sqrt 2 a)}}{{b{q_r}R{\rho _1}}}, \notag\\
{\gamma _2} &= 2J_0^2(\sqrt 2 a) + \frac{{{V_c}{J_0}(\sqrt 2 a)}}{{b{q_r}R{\rho _2}}}, \notag\\
{\gamma _3} &= \frac{{{J_0}(2\sqrt 2 a) + {J_0}(2a) - {\gamma _2}}}{{{J_0}(2\sqrt 2 a) - 1}}, \notag\\
{\rho _1} &= 2{J_0}^2(\sqrt 2 a) - \frac{1}{2}\left[ {{J_0}(2\sqrt 2 a) + {J_0}(2a) + 1} \right], \notag\\
{\rho _2} &= \frac{{\sqrt 2 b\left[ {1 - {J_0}(2\sqrt 2 a)} \right]}}{{4{c_\theta }{J_1}(\sqrt 2 a)}}, \notag\\
{\mu _0} &= \arccos \left( { - \sqrt {\frac{1}{{2{\gamma _3}}}} } \right), \notag
\end{align}
\begin{align}
{e_{1}} &=  - \frac{{V_c^2J_0^2(\sqrt 2 a)}}{{{q_r}{b^2}{R^2}\rho _1^2}} + 2\frac{{{V_c}J_0^2(\sqrt 2 a)}}{{b{\rho _1}}}, \notag\\
{e_{2}} &=  - 2{q_r}{\gamma _3}{\rho _2}^2 - 2{q_r}R{\rho _2}{J_0}(\sqrt 2 a). \notag
\end{align}

The average system (\ref{eq:ave1})-(\ref{eq:ave5}) may converge to one of the four equilibria (\ref{eq:eq1})-(\ref{eq:eq4}) under different parameters and initial conditions. Each equilibrium corresponds to a torus in the vicinity of  the source. Around equilibrium (\ref{eq:eq1}) and equilibrium (\ref{eq:eq2}), the vehicle converges to a torus with its average heading directly towards or away from the source. Around equilibrium (\ref{eq:eq3}) and equilibrium (\ref{eq:eq4}) the vehicle revolves around the source clockwise or counterclockwise and its average heading is more outward than inward. The ${\tilde r_c^{\rm{ave}}}$ should be real and positive as it represents the average distance between the vehicle center and the source. Note that ${\alpha ^{*{\rm{ave}}}},{{\hat \alpha }^{{\rm{ave}}}}$ are equal to zero for all the equilibria of the average system, thus all the tori nearly reduce to annuluses. The values of ${\tilde \theta }^{{\rm{ave}}}$ for different average equilibria are quite different, whilts in different motion patterns in the tori. We shall formalize these phenomena in the sequel and illustrate in the simulations.

Let ${J^{eq1}}$, ${J^{eq2}}$, ${J^{eq3}}$ and ${J^{eq4}}$ denote the Jacobians of the average system (\ref{eq:ave1})-(\ref{eq:ave5}) at equilibria (\ref{eq:eq1})-(\ref{eq:eq4}) respectively. It is obvious that if all the roots of the characteristic equation for a Jacobian have negative real parts,  the  corresponding equilibrium is an exponentially stable equilibrium of the average system. Noting from (\ref{eq:j1}) and (\ref{eq:j2}) that the characteristic equations of ${J^{eq1}}$ and ${J^{eq2}}$ are the same, we conclude that equilibrium (\ref{eq:eq1}) and equilibrium (\ref{eq:eq2}) have the same stability conditions. This property also holds for equilibrium (\ref{eq:eq3}) and equilibrium (\ref{eq:eq4}). Applying Theorem  in \cite{Khalil2002}, we derive the following two propositions.

\begin{proposition}
Consider the system (\ref{eq:sys1})-(\ref{eq:sys6}) with positive parameters $a$, $c_\alpha$, $c_\theta$, $b$, $h$ and $V_c$. Suppose that these parameters are appropriately chosen so that the Jacobian ${J^{eq1}}$ is Hurwitz. For sufficiently large $\omega$, if the initial conditions ${r_c}(0),\theta (0),\alpha (0),e(0)$ are such that the following quantities are sufficiently small
\[\begin{array}{l}
\big| {\left| {{r_c}(0) - r^*} \right| - \left| {{\gamma _1}} \right|} \big|, \left| {\alpha (0)} \right|, \left| {e(0) - {q_r}{R^2} - {e_1}} \right|,\\
\rm{and \quad} \left| {\theta (0) - \arctan \frac{{{y_c} - {y^*}}}{{{x_c} - {x^*}}} - \frac{\pi }{2} + {\mathop{\rm sgn}} ({\gamma_1}) \times \frac{\pi }{2}} \right|,
\end{array}\]
where the ${\mathop{\rm sgn}} ( \cdot )$ is a standard sign function, then the trajectory of the vehicle center $r_c(t)$ exponentially converges to, and remains in the torus
\begin{align}
& \left| \alpha  \right|\le O(1/\omega ), \notag\\
{\left|\gamma _1\right|} - O(1/\omega ) &\le \left| {{r_c} - {r^*}} \right| \le \left|\gamma _1\right| + O(1/\omega ). \notag
\end{align}
\end{proposition}
\begin{IEEEproof}
The system (\ref{eq:sys1})-(\ref{eq:sys6}) is equivalent to the error system system (\ref{eq:es1})-(\ref{eq:es5}), whose solution can be approximated by the solution of the corresponding ``average system" according to the averaging theory in \cite{Khalil2002}. If the Jacobian ${J^{eq1}}$ is Hurwitz, then both equilibria (\ref{eq:eq1}) and (\ref{eq:eq2}) are exponentially stable of the average system since the characteristic equations of ${J^{eq1}}$ and ${J^{eq2}}$ are the same. By Theorem 10.4 in \cite{Khalil2002}, we conclude that the error system (\ref{eq:es1})-(\ref{eq:es5}) has two distinct, exponentially stable periodic solutions within $O(1/\omega)$ of the equilibria (\ref{eq:eq1}) and (\ref{eq:eq2}).
\end{IEEEproof}

\begin{proposition}
Consider the system (\ref{eq:sys1})-(\ref{eq:sys6}) with positive parameters $a$, $c_\alpha$, $c_\theta$, $b$, $h$ and $V_c$. Suppose that these parameters are appropriately chosen so that the Jacobian ${J^{eq3}}$ is Hurwitz. For sufficiently large $\omega$, if the initial conditions ${r_c}(0),\theta (0),\alpha (0),e(0)$ are such that the following quantities are sufficiently small
\[\begin{array}{l}
\left| {\left| {{r_c}(0) - r^*} \right| - {{\rho _2}\sqrt {2{\gamma _3}} }} \right|,\left| {\alpha (0)} \right|,\left| {e(0) - {q_r}{R^2} - {e_2}} \right|, \\
\rm{and \quad either \quad} \left| {\theta (0)} - \arctan \frac{{{y_c} - {y^*}}}{{{x_c} - {x^*}}} - {{\mu _0}} \right| \\
\rm{or \quad} \left| {\theta (0)}  - \arctan \frac{{{y_c} - {y^*}}}{{{x_c} - {x^*}}} + {{\mu _0}} \right|,
\end{array}\]
then the trajectory of the vehicle center $r_c(t)$ exponentially converges to, and remains in the torus
\begin{align}
& \left| \alpha  \right|\le O(1/\omega ), \notag\\
{\rho _2}\sqrt {2{\gamma _3}} - O(1/\omega ) &\le \left| {{r_c} - {r^*}} \right| \le {\rho _2}\sqrt {2{\gamma _3}} + O(1/\omega ). \notag
\end{align}
\end{proposition}

Here we do not give the proof of Proposition 2 since it is essentially a repetition of the proof of Proposition 1. In both propositions, the vehicle center converges to a torus around the source. In Proposition 1, the limit of the vehicle's heading is either directly towards or away from the source on average. In this case the vehicle stays in the torus as if it comes to a full stop, which is quite different from the result of Cochran's work \cite{Cochran2009a}. In Proposition 2, the vehicle drifts clockwise or counterclockwise in the torus and the limit of its average heading is more outward than inward.

\begin{remark}
It should be noted that we have assumed the distribution of the signal field can be approximated by $$J = {f^*} - {q_r}{\left| {{r_s} - {r^*}} \right|^2}$$ in the stability analysis. This implies that the proposed control scheme works for signal fields satisfying the following two requirements: (a) the signal strength depends only on the distance between the vehicle and the source, i.e., $J = f(d),\; d=|r_s-r^*|$; (b) $f(d)$ is strictly concave in the distance $d$. However, the simulation results in Section \ref{sec_simulation} illustrate that the control scheme also works for some other unknown forms of the signal fields.
\end{remark}

\section{Parameter Selection}
In this section we further study the stability conditions for Proposition 1, and provide a rule on the selection of the parameters in Proposition 1. The stability conditions for Proposition 2 can be similarly derived. However, their forms are fairly complex and lengthy, we only give a brief discussion in Appendix \ref{appendix_stability}.

We first note that the Jacobians of the average system (\ref{eq:ave1})-(\ref{eq:ave5}) at equilibrium (\ref{eq:eq1}) and equilibrium (\ref{eq:eq2}) are given in the following form
\begin{equation}
{J^{eq1}} = \frac{1}{\omega }\left[ {\begin{matrix}
{m_{11}}&0&0&0&{m_{15}}\\
0&{m_{22}}&{m_{23}}&0&0\\
0&{m_{32}}&{m_{33}}&0&0\\
0&0&0&{{m_{44}}}&0\\
{m_{51}}&0&0&0&{-h}
\end{matrix}} \right],
\label{eq:j1}
\end{equation}
\begin{equation}
{J^{eq2}} = \frac{1}{\omega }\left[ {\begin{matrix}
{m_{11}}&0&0&0&{-m_{15}}\\
0&{m_{22}}&{-m_{23}}&0&0\\
0&{-m_{32}}&{m_{33}}&0&0\\
0&0&0&{{m_{44}}}&0\\
{-m_{51}}&0&0&0&{-h}
\end{matrix}} \right],
\label{eq:j2}
\end{equation}
where
\begin{align}
{m_{11}} &= \frac{{2{V_c}{J_0}^2(\sqrt 2 a)}}{{R{\rho _1}}} - \frac{1}{2}b{q_r}R\left[ {{J_0}(2\sqrt 2 a) + 2{J_0}(2a) + 1} \right],\notag\\
{m_{15}} &= b{J_0}(\sqrt 2 a),\notag\\
{m_{22}} &= \frac{1}{2}b{q_r}R\left[ {3{J_0}(2a) - 2} \right],\notag\\
{m_{23}} &= b{q_r}R\frac{{{J_0}(a)}}{{2{J_0}(\sqrt 2 a)}}\left[ {{J_0}(2\sqrt 2 a) + {J_0}(2a) + 1} \right] \notag\\
&\quad - 2b{q_r}R{J_0}(\sqrt 5 a),\notag\\
{m_{32}} &= 2{c_\alpha }\frac{{{V_c}{J_0}(\sqrt 2 a)}}{{b{\rho _1}}}{J_1}(a),\notag\\
{m_{33}} &=  - \sqrt 2 {c_\alpha }\frac{{{V_c}{J_0}(\sqrt 2 a)}}{{b{\rho _1}}}{J_1}(\sqrt 2 a),\notag\\
{m_{44}} &=  - \sqrt 2 {c_\theta }\frac{{{V_c}{J_0}(\sqrt 2 a)}}{{b{\rho _1}}}{J_1}(\sqrt 2 a) \notag\\
&\quad + \frac{1}{2}b{q_r}R\left[ {{J_0}(2\sqrt 2 a) + {J_0}(2a) - 1} \right],\notag\\
{m_{51}} &=  - 2h\frac{{{V_c}{J_0}(\sqrt 2 a)}}{{bR{\rho _1}}} + 2h{q_r}R{J_0}(\sqrt 2 a).\notag
\end{align}

Due to the property of determinant \cite{horn2012matrix}, ${J^{eq1}}$ has the same eigenvalues as the following block diagonal matrix
\begin{equation}
\frac{1}{\omega }\cdot\rm{diag} \left\{
\left[ \begin{matrix}
{m_{22}}&{m_{23}}\\
{m_{32}}&{m_{33}}
\end{matrix}
\right],\;
\begin{bmatrix}
m_{44}
\end{bmatrix},\;
\left[ \begin{matrix}
{m_{11}}&{m_{15}}\\
{m_{51}}&{-h}
\end{matrix}
\right]
\right\}.
\label{jocabian}
\end{equation}
It is clear that the characteristic equation for matrix (\ref{jocabian}) is the product of the characteristic equations of the three blocks. Then we can easily calculate the characteristic equation for $J^{eq1}$, which is given by
\begin{align}
0 &= \left[ {{{(\omega s)}^2} - ({m_{22}} + {m_{33}})\omega s + {m_{22}}{m_{33}} - {m_{23}}{m_{32}}} \right] \notag\\
&\quad\times \left( {\omega s - {m_{44}}} \right) \notag\\
&\quad\times\left[ {{{(\omega s)}^2} + (h - {m_{11}})\omega s - {m_{11}}h - {m_{15}}{m_{51}}} \right]. \label{eq:characteristic}
\end{align}

Note that the characteristic equation for ${J^{eq2}}$ is also equal to (\ref{eq:characteristic}), thus equilibrium (\ref{eq:eq1}) and equilibrium (\ref{eq:eq2}) have the same stability conditions.

To guarantee that all the roots of characteristic equation (\ref{eq:characteristic}) have negative real parts, we invoke the Routh-Hurwitz test \cite{routh1877treatise} and result in the following conditions
\begin{align}
{m_{22}} + {m_{33}} < 0,\\
{m_{22}}{m_{33}} - {m_{23}}{m_{32}} > 0,\\
{m_{44}} < 0,\\
h - {m_{11}} > 0,\\
 - {m_{11}}h - {m_{15}}{m_{51}} > 0.
\end{align}
Substituting the definitions of ${m_{11}},\cdots,{m_{51}}$ into the above inequalities, we derive the following stability conditions
\begin{align}
{b^2}{q_r}R{\phi _1} < 2\sqrt 2 {c_\alpha }{V_c}\frac{{{J_0}(\sqrt 2 a)}}{{{\rho _1}}}{J_1}(\sqrt 2 a),\label{eq:cc1}\\
{b^2}{q_r}R{\phi _2} < 2\sqrt 2 {c_\theta }{V_c}\frac{{{J_0}(\sqrt 2 a)}}{{{\rho _1}}}{J_1}(\sqrt 2 a),\label{eq:cc2}\\
2{V_c}\frac{{{J_0}^2(\sqrt 2 a)}}{{{\rho _1}}} < hR + \frac{1}{2}b{q_r}{R^2}{\phi _4},\label{eq:cc3}\\
\frac{{{J_0}(\sqrt 2 a)}}{{{\rho _1}}}\left[ {\frac{{\sqrt 2 }}{2}{\phi _1}{J_1}(\sqrt 2 a) + {\phi _3}{J_1}(a)} \right] < 0,\label{eq:cc4}\\
4{J_0}^2(\sqrt 2 a) - {\phi _4} < 0,\label{eq:cc5}
\end{align}
where
\begin{align}
{\phi _1} &= 3{J_0}(2a) - 2,\notag\\
{\phi _2} &= {J_0}(2\sqrt 2 a) + {J_0}(2a) - 1,\notag\\
{\phi _3} &= \frac{{{J_0}(a)}}{{{J_0}(\sqrt 2 a)}}\left( {{\phi _2} + 2} \right) - 4{J_0}(\sqrt 5 a),\notag\\
{\phi _4} &= {J_0}(2\sqrt 2 a) + 2{J_0}(2a) + 1. \notag
\end{align}

Note that inequalities (\ref{eq:cc4}) and (\ref{eq:cc5}) depend only on the parameter $a$. We restrict the parameter $a$ to the union of two intervals $\mathbf{S}_a^1$ and $\mathbf{S}_a^2$, which are defined by $\mathbf{S}_a^1 = \left[ {1.25, 1.65} \right]$ and $\mathbf{S}_a^2 = \left[ {1.75, 2.5} \right]$. The reason is that the constraint guarantees the validity of inequalities (\ref{eq:cc4}) and (\ref{eq:cc5}) and also brings convenience for the design of the other parameters. Under this constraint, we have that $\phi_1<0$, $\phi_2<0$, ${\rho _1}<0$, ${J_1}(\sqrt 2 a)>0$, and
\[\left\{ \begin{array}{l}
{J_0}(\sqrt 2 a)> 0, \quad \rm{if} \; a\in \mathbf{S}_a^1,\\
{J_0}(\sqrt 2 a) < 0, \quad \rm{if} \; a\in \mathbf{S}_a^2.
\end{array} \right.\]

Now let us consider inequalities (\ref{eq:cc1})-(\ref{eq:cc3}). Observe that the right hand side of inequality (\ref{eq:cc3}) is positive since $\phi_4$ is positive for all positive $a$. The left hand side of inequality (\ref{eq:cc3}) is negative with a negative ${\rho _1}$. Thus inequality (\ref{eq:cc3}) holds for all $a\in \mathbf{S}_a^1 \cup \mathbf{S}_a^2$. When $a\in \mathbf{S}_a^2$, the inequalities (\ref{eq:cc1}) and (\ref{eq:cc2}) hold since their left hand side is negative while their right hand side is positive. When $a\in \mathbf{S}_a^1$, we need the following conditions
\begin{align}
{V_c} < \frac{{\sqrt 2 {b^2}{q_r}R{\phi _1}{\rho _1}}}{{4{c_\alpha }{J_0}(\sqrt 2 a){J_1}(\sqrt 2 a)}},\\
{V_c} < \frac{{\sqrt 2 {b^2}{q_r}R{\phi _2}{\rho _1}}}{{4{c_\theta }{J_0}(\sqrt 2 a){J_1}(\sqrt 2 a)}},
\end{align}
to satisfy the inequalities (\ref{eq:cc1}) and (\ref{eq:cc2}).

\begin{figure*}[!t]
  \centering
  \includegraphics[width=0.9\linewidth]{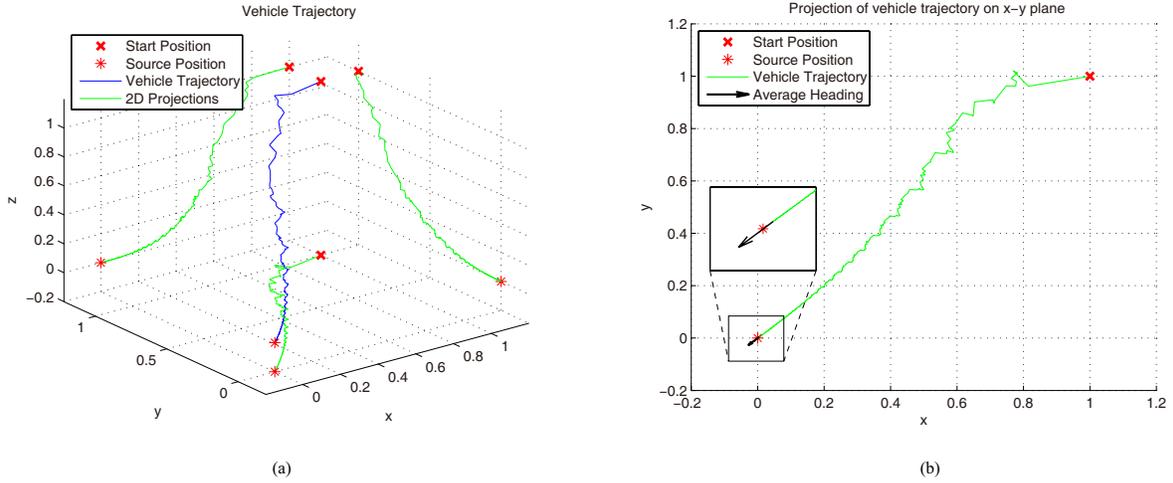}
  \caption{Simulation results for Corollary 1: (a) Vehicle trajectory; (b) Projection of vehicle trajectory on $x-y$ plane.}
  \label{figeq1}
\end{figure*}

Summarizing the above, we conclude that the local exponential convergence to equilibria (\ref{eq:eq1}) and (\ref{eq:eq2}) can be guaranteed by selecting appropriate parameters $a$ and $V_c$ as in the following two corollaries:

\begin{corollary}
Consider the system (\ref{eq:sys1})-(\ref{eq:sys6}) with positive parameters $a$, $c_\alpha$, $c_\theta$, $b$, $h$ and $V_c$. Let the parameter $a\in \left[ {1.75, 2.5} \right]$, then for sufficiently large $\omega$, if the initial conditions ${r_c}(0),\theta (0),\alpha (0),e(0)$ are such that the following quantities are sufficiently small
\[\begin{array}{l}
\big| {\left| {{r_c}(0) - r^*} \right| - {\gamma _1}} \big|, \left| {\alpha (0)} \right|, \left| {e(0) - {q_r}{R^2} - {e_1}} \right|,\\
\rm{and \quad} \left| {\theta (0) - \arctan \frac{{{y_c} - {y^*}}}{{{x_c} - {x^*}}}} \right|,
\end{array}\]
then the trajectory of the vehicle center $r_c(t)$ exponentially converges to, and remains in the torus
\begin{align}
& \left| \alpha  \right|\le O(1/\omega ), \notag\\
{\gamma _1} - O(1/\omega ) &\le \left| {{r_c} - {r^*}} \right| \le {\gamma _1} + O(1/\omega ). \notag
\end{align}
\end{corollary}
\begin{corollary}
Consider the system (\ref{eq:sys1})-(\ref{eq:sys6}) with positive parameters $a$, $c_\alpha$, $c_\theta$, $b$, $h$ and $V_c$. Let the parameter $a\in \left[ {1.25, 1.65} \right]$ and the parameter $V_c$ be chosen such that $V_c<\overline{V_c}$, where
\begin{equation}
\overline {{V_c}} \buildrel \Delta \over =
\frac{{\sqrt 2 {b^2}{q_r}R}}{{4{J_0}(\sqrt 2 a){J_1}(\sqrt 2 a)}} \times \min \left\{ \frac{\phi_1\rho_1}{c_\alpha}, \frac{\phi_2\rho_1}{c_\theta} \right\},
\end{equation}
then for sufficiently large $\omega$, if the initial conditions ${r_c}(0),\theta (0),\alpha (0),e(0)$ are such that the following quantities are sufficiently small
\[\begin{array}{l}
\big| {\left| {{r_c}(0) - r^*} \right| + {\gamma _1}} \big|, \left| {\alpha (0)} \right|, \left| {e(0) - {q_r}{R^2} - {e_1}} \right|,\\
\rm{and \quad} \left| {\theta (0) - \arctan \frac{{{y_c} - {y^*}}}{{{x_c} - {x^*}}} - \pi} \right|,
\end{array}\]
then the trajectory of the vehicle center $r_c(t)$ exponentially converges to, and remains in the torus
\begin{align}
& \left| \alpha  \right|\le O(1/\omega ), \notag\\
-{\gamma _1} - O(1/\omega ) &\le \left| {{r_c} - {r^*}} \right| \le -{\gamma _1} + O(1/\omega ). \notag
\end{align}
\end{corollary}

The above two corollaries provide a rule on the parameter configuration of the control scheme for the static source that produces a signal field with spherical level sets. In the two corollaries, the vehicle center locally exponentially converges to, and remains in the torus represented by equilibrium (\ref{eq:eq1}) or equilibrium  (\ref{eq:eq2}). Note that the conditions in the corollaries are usually stricter than the stability conditions (\ref{eq:cc1})-(\ref{eq:cc5}). In fact, we can find many arrays of parameters that satisfy the stability conditions (\ref{eq:cc1})-(\ref{eq:cc5}) but violate the conditions in Corollary 1 and Corollary 2. It is interesting that the biased forward velocity $V_c$ in Corollary 2 must be small enough while it can be any positive value in Corollary 1. This distinct feature is different from the 2-D case in \cite{Ghods2010}, which requires a small $V_c$ to actuate the vehicle to ``stop" near the source. However, we still suggest a relatively small $V_c$ for Corollary 1 since a smaller $V_c$ results in a closer distance to the source.
\begin{remark}
In this section, the intervals [1.75, 2.5] and [1.25, 1.65] are suggested for the perturbation amplitude since they guarantee the validity of inequalities (43) and (44) and also bring convenience for the design of the other parameters. Besides, a larger perturbation amplitude implies more resistant to the noise. In Cochran's work \cite{Cochran2009a}, such a large perturbation amplitude would lead to rambling around the source, as a constant forward velocity and a large perturbation amplitude make the vehicle move fast. When around the source, the vehicle moves too fast to estimate the gradient precisely. In our control scheme, the forward velocity is designed to be positively correlated to the output of the washout filter. Such a tunable forward velocity can avoid ¡°too fast¡± movements around the source. Thus even the perturbation amplitude is chosen within [1.75, 2.5] and [1.25, 1.65], the vehicle can estimate the gradient correctly and settle down near the source.
\end{remark}

The derivation for stability conditions for Proposition 2 is provided in Appendix \ref{appendix_stability}.

\section{Simulation}
\label{sec_simulation}
Due to different initial conditions, the vehicle may converge to different regions around the source under different scenarios. In this section we present simulation results to illustrate the behaviors described in Corollary 1, Corollary 2 and Proposition 2. We also consider locating a source which creates a non-quadratic signal field.

\subsection{Signal fields with spherical and elliptical level sets}

\begin{figure}[!t]
  \centering
  \includegraphics[width=0.9\hsize]{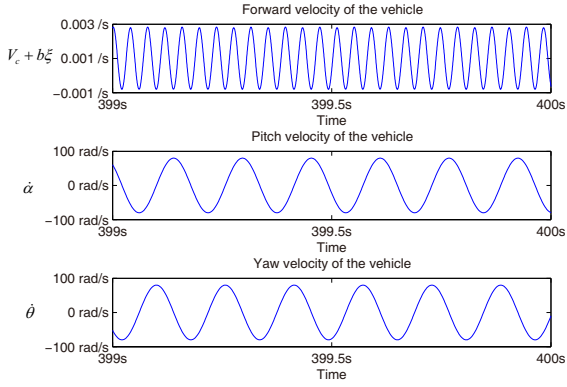}
  \caption{Forward and angular velocities of the vehicle.}
  \label{figeq1V}
\end{figure}

Fig.~\ref{figeq1} illustrates the behavior dictated by Corollary 1. The map parameters are set as $f^* = 1$, $r^* = (0,0,0)$ and $q_r = 1$, and the initial conditions of the vehicle are set as $r_c(0) = (1,1,1)$, $\alpha(0) = -\pi/2$ and $\theta(0) = -\pi/2$. The controller parameters are chosen as $\omega = 40$, $a = 2$, $c_\alpha = 100$, $c_\theta=100$, $b = 5$, $h = 10$ and $V_c = 0.001$. Fig.~\ref{figeq1} (a) shows the trajectory of the vehicle converging to an attractor very close to the source. Instead of drifting in a torus like in \cite{Cochran2009a}, the vehicle slows down nearly to a stop when it is in the torus, which is almost in a horizontal plane, as the average azimuthal angle between the vehicle and the source is close to zero. We can notice more details in Fig.~\ref{figeq1} (b), which describes the projection of vehicle trajectory on $x-y$ plane. The limit of the average heading of the vehicle is directly towards the source, which coincides with the theoretical result in Corollary 1. Fig.~\ref{figeq1V} depicts the forward and angular velocities of the vehicle in the torus.

\begin{figure}[!t]
  \centering
  \includegraphics[width=0.9\hsize]{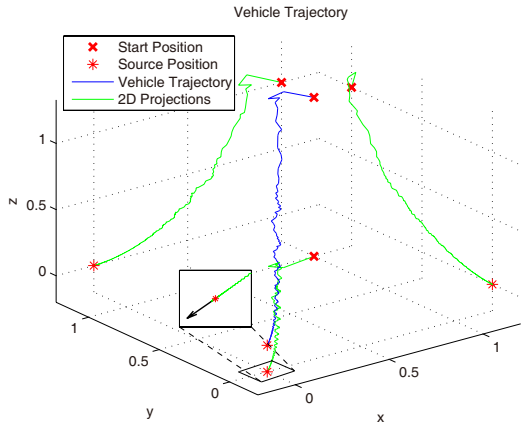}
  \caption{Vehicle trajectory illustrating Corollary 2.}
  \label{figeq2}
\end{figure}

Fig.~\ref{figeq2} illustrates the behavior dictated by Corollary 2. The map parameters, initial conditions of the vehicle, and controller parameters are chosen to be the same as those in Fig.~\ref{figeq1} except $a = 1.5$. As in Fig.~\ref{figeq1}, the vehicle converges to an approximately planar near to the source. In this case the vehicle overshoots the source and comes to a ``stop". The limit of the average heading of the vehicle is directly away from the source, as dictated by the average equilibrium (\ref{eq:eq2}).

Note that the vehicle does not strictly come to a full stop in the two simulations though its stop seems evident from Fig.~\ref{figeq1} and Fig.~\ref{figeq2}. In fact, the forward velocity of the vehicle oscillates around the biased forward velocity $V_c$ and the yaw and pitch velocities oscillate around zero, as shown in Fig.~\ref{figeq1V}. When the vehicle settles stably in the torus, it still remains moving within a small area.

\begin{figure}[!t]
  \centering
  \includegraphics[width=0.9\hsize]{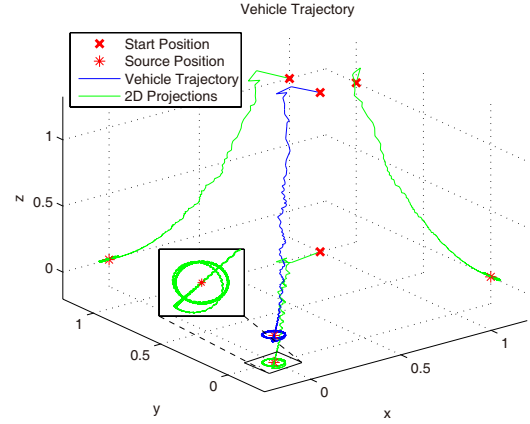}
  \caption{Vehicle trajectory illustrating Proposition 2.}
  \label{figeq3}
\end{figure}

Fig.~\ref{figeq3} illustrates the behavior dictated by Proposition 2. The map parameters, initial conditions of the vehicle, and controller parameters are chosen to be the same as those in Fig.~\ref{figeq2} except $V_c = 0.1$. The trajectory before entering a torus is similar to the one in Fig.~\ref{figeq2}. However, the vehicle turns around and keeps moving with a relatively large forward velocity after it overshoots the source. After several times of adjustment of direction, it eventually drifts in a planar torus, as shown in Fig.~\ref{figeq3}.

\begin{figure}[!t]
  \centering
  \includegraphics[width=0.9\hsize]{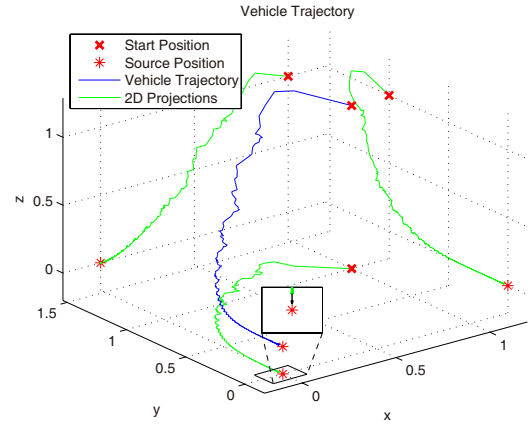}
  \caption{Vehicle locates a source which creates a signal field with elliptical level sets.}
  \label{figeq5}
\end{figure}

Our control scheme also allows the vehicle to seek a source which creates a signal field with elliptical level sets, as shown in Fig.~\ref{figeq5}. We now assume the nonlinear map takes the form
$$f({r_s}) = 1 - 2x_s^2 - 0.5y_s^2 - z_s^2,$$
where ${r_s} = {\left[{x_s}, {y_s},{z_s}\right]^T}$. The initial conditions of the vehicle and controller parameters are chosen to be the same as those in Fig.~\ref{figeq1}. The vehicle converges to a region near the source with its average heading directly towards the source, which is similar to the result in Fig.~\ref{figeq1}.
\subsection{Non-quadratic and non-concave signal fields}
\begin{figure}[!t]
  \centering
  \includegraphics[width=0.9\hsize]{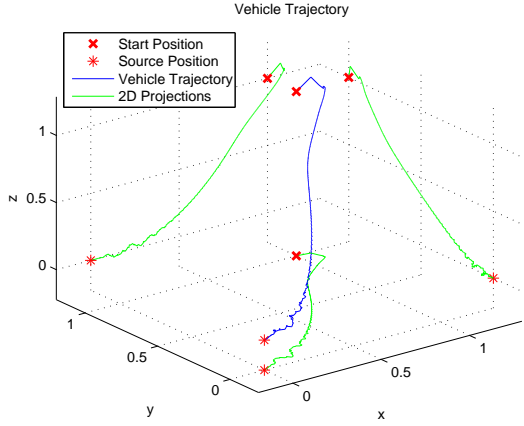}
  \caption{Vehicle approaches an acoustic source.}
  \label{figeq6}
\end{figure}

\begin{figure}[!t]
  \centering
  \includegraphics[width=0.9\hsize]{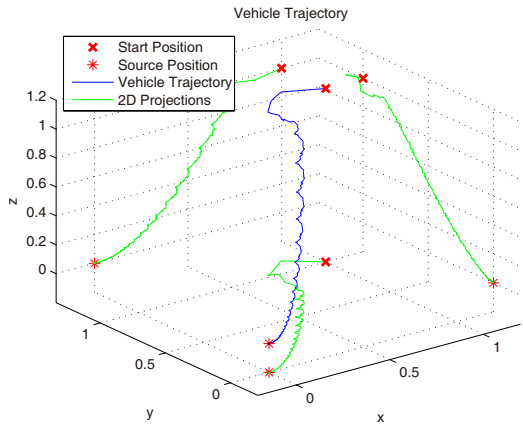}
  \caption{Vehicle approaches a source whose signal map is a Rosenbrock function.}
  \label{figeq7}
\end{figure}

For some signal fields with non-quadratic maps, our control scheme also exhibits abilities to seek the sources, as shown in Fig.~\ref{figeq6} and Fig.~\ref{figeq7}. In Fig.~\ref{figeq6} we consider an acoustic source with unit emitting power. Then, the signal distribution takes the form
$$J = f({r_s}) = \frac{1}{4\pi\left| r_s\right|^2}.$$ Considering the signal strength is too large near the source, we replace $J$ by $J'$, which takes the form
$$J' = -\exp(-J) = -\exp\left(-\frac{1}{4\pi\left| r_s\right|^2}\right).$$
In Fig.~\ref{figeq7} we assume the distribution of the signal field is an Rosenbrock function, which takes the form
$$J = -x_s^2-(y_s-x_s^2)^2-y_s^2-(z_s-y_s^2)^2.$$
The Rosenbrock function is obviously a non-concave function and has an isolated maximum at (0,0,0). In both simulations, the initial conditions of the vehicle and controller parameters are chosen to be the same as those in Fig.~\ref{figeq1}. Simulations show the vehicle can well approach the source, as depicted in Fig.~\ref{figeq6} and Fig.~\ref{figeq7}.


\section{Conclusion}

We have studied the nonholonomic source seeking problem in the 3-D workspace by proposing a new control strategy, which extends the work in the 2-D workspace \cite{Ghods2010} from two dimensions to three dimensions, and proved the local exponential convergence of the closed-loop system for a static source that produces a signal field with spherical level sets. Under different parameters the vehicle may virtually ``stop" or revolve around the source. We  also provided a rule on the parameter selection for the control scheme. The theoretical results were validated through simulations under different scenarios.

\appendices
\section{Computation of the Equilibria}
\label{appendix_equlibira}

By setting the right hand side of the average error system (\ref{eq:ave1})-(\ref{eq:ave5}) to be zero, we have
\begin{align}
0 &= {{(b{q_r}{{(\tilde r_c^{{\rm{ave}}})}^2} + b{{\hat e}^{{\rm{ave}}}} - {V_c})}}\xi _c^{{\rm{ave}}} - {{2b{q_r}R\tilde r_c^{{\rm{ave}}}}}\xi _c^{2{\rm{ave}}}, \label{ap1} \\
0 &= \frac{{(b{q_r}{{(\tilde r_c^{{\rm{ave}}})}^2} + b{{\hat e}^{{\rm{ave}}}} - {V_c})}}{{ \tilde r_c^{{\rm{ave}}}}}{\xi ^{\genfrac{}{}{0pt}{}{\scriptstyle\alpha\hfill}{\scriptstyle{\rm{ave}}}}}
- {2b{q_r}R}\xi _c^{\genfrac{}{}{0pt}{}{\scriptstyle\alpha\hfill}{\scriptstyle{\rm{ave}}}}, \label{ap2}\\
0 &= \xi _c^{\genfrac{}{}{0pt}{}{\scriptstyle\sin}{\scriptstyle{\rm{ave}}}}, \label{ap3}\\
0 &= {2{c_\theta }{q_r}R\tilde r_c^{{\rm{ave}}}}\xi _c^{\genfrac{}{}{0pt}{}{\scriptstyle\cos}{\scriptstyle{\rm{ave}}}} + \frac{{2b{q_r}R}}{{ \cos ({\alpha ^{*{\rm{ave}}}})}}\xi _c^{\genfrac{}{}{0pt}{}{\scriptstyle\cos\sin}{\scriptstyle{\rm{ave}\hfill}}} \notag \\
& \quad + \frac{{({V_c} - b{q_r}{{(\tilde r_c^{{\rm{ave}}})}^2} - b{{\hat e}^{{\rm{ave}}}})}}{{\tilde r_c^{{\rm{ave}}}\cos ({\alpha ^{*{\rm{ave}}}})}}{\xi ^{\genfrac{}{}{0pt}{}{\scriptstyle\cos\sin}{\scriptstyle{\rm{ave}\hfill}}}}, \label{ap4}\\
0 &=  - {q_r}{(\tilde r_c^{{\rm{ave}}})^2} - {{\hat e}^{{\rm{ave}}}} + {2{q_r}R}\tilde r_c^{{\rm{ave}}}\xi _c^{{\rm{ave}}}. \label{ap5}
\end{align}

We first note that when ${\alpha ^*}^{{\rm{ave}}} = 0$ and ${\hat \alpha }^{{\rm{ave}}} = 0$, the equations (\ref{ap2}) and (\ref{ap3}) hold since ${\xi ^{\genfrac{}{}{0pt}{}{\scriptstyle\alpha\hfill}{\scriptstyle{\rm{ave}}}}} = 0$, $\xi _c^{\genfrac{}{}{0pt}{}{\scriptstyle\alpha\hfill}{\scriptstyle{\rm{ave}}}} = 0$ and $\xi _c^{\genfrac{}{}{0pt}{}{\scriptstyle\sin}{\scriptstyle{\rm{ave}}}} = 0$. Then we only need to consider the equations (\ref{ap1}), (\ref{ap4}) and (\ref{ap5}), which take the forms
\begin{align}
0 &= -\eta_1{J_0}(\sqrt 2 a)\cos ({{\tilde \theta }^{^{{\rm{ave}}}}}) - \frac{1}{2}b{q_r}R\tilde r_c^{{\rm{ave}}}\eta_2, \label{nap1}\\
0 &=  - {\sqrt 2}{c_\theta }{q_r}R\tilde r_c^{{\rm{ave}}}{{J_1}(\sqrt 2 a)}\sin ({{\tilde \theta }^{{\rm{ave}}}})\notag\\
&\quad + \frac{\eta_1}{{\tilde r_c^{{\rm{ave}}}}}{J_0}(\sqrt 2 a)\sin ({{\tilde \theta }^{{\rm{ave}}}}) \notag\\
&\quad + \frac{1}{2}b{q_r}R\left[ {{J_0}(2\sqrt 2 a) + {J_0}(2a)} \right]\sin (2{{\tilde \theta }^{{\rm{ave}}}}), \label{nap2}\\
{{\hat e}^{{\rm{ave}}}} &=  - {q_r}{(\tilde r_c^{{\rm{ave}}})^2} + 2{q_r}R\tilde r_c^{{\rm{ave}}}{J_0}(\sqrt 2 a)\cos ({{\tilde \theta }^{^{{\rm{ave}}}}}),\label{nap3}
\end{align}
where
\begin{align}
\eta_1 &= {V_c} - b{q_r}{{(\tilde r_c^{{\rm{ave}}})}^2} - b{{\hat e}^{{\rm{ave}}}}, \notag \\
\eta_2 &= {\left[ {{J_0}(2\sqrt 2 a) + {J_0}(2a)} \right]\cos (2{{\tilde \theta }^{{\rm{ave}}}}) + {J_0}(2a) + 1}. \notag
\end{align}
When $\sin({{\tilde \theta }^{{\rm{ave}}}} )= 0$, the equation (\ref{nap2}) holds and we can easily derive equilibria (\ref{eq:eq1}) and (\ref{eq:eq2}). Otherwise we rewrite the equations (\ref{nap1}) and equation (\ref{nap2}) as follows
\begin{align}
\frac{\eta_1}{{\tilde r_c^{{\rm{ave}}}}} &= -\frac{{b{q_r}R\eta_2 }}{{2{J_0}(\sqrt 2 a)\cos ({{\tilde \theta }^{^{{\rm{ave}}}}})}}, \label{fap1}\\
0 &=  - {\sqrt 2}{c_\theta }{q_r}R\tilde r_c^{{\rm{ave}}}{{J_1}(\sqrt 2 a)} + \frac{\eta_1}{{\tilde r_c^{{\rm{ave}}}}}{J_0}(\sqrt 2 a) \notag\\
&\quad + b{q_r}R\left[ {{J_0}(2\sqrt 2 a) + {J_0}(2a)} \right]\cos({{\tilde \theta }^{{\rm{ave}}}}). \label{fap2}
\end{align}
Substituting (\ref{fap1}) into (\ref{fap2}), we derive
\begin{equation}
{\tilde r_c^{{\rm{ave}}}} = -\frac{\rho_2}{\cos({{\tilde \theta }^{{\rm{ave}}}})}. \label{fap3}
\end{equation}
Combining (\ref{nap1}) and (\ref{fap3}), we figure out ${\tilde r_c^{{\rm{ave}}}}$. Then we obtain equilibria (\ref{eq:eq3}) and (\ref{eq:eq4}).

\section{Stability Conditions for Proposition 2}
\label{appendix_stability}

The Jacobians of the average system (\ref{eq:ave1})-(\ref{eq:ave5}) at equilibrium (\ref{eq:eq3}) and equilibrium (\ref{eq:eq4}) are
\begin{equation}
{J^{eq3}} = \frac{1}{\omega }\left[ {\begin{matrix}
{l_{11}}&0&0&{l_{14}}&{l_{15}}\\
0&{l_{22}}&{l_{23}}&0&0\\
0&{l_{32}}&{l_{33}}&0&0\\
{{l_{41}}}&0&0&{{l_{44}}}&{{l_{45}}}\\
{l_{51}}&0&0&{{l_{54}}}&{-h}
\end{matrix}} \right],\notag
\end{equation}
\begin{equation}
{J^{eq4}} = \frac{1}{\omega }\left[ {\begin{matrix}
{l_{11}}&0&0&{-l_{14}}&{l_{15}}\\
0&{l_{22}}&{l_{23}}&0&0\\
0&{l_{32}}&{l_{33}}&0&0\\
{{-l_{41}}}&0&0&{{l_{44}}}&{-{l_{45}}}\\
{l_{51}}&0&0&{-{l_{54}}}&{-h}
\end{matrix}} \right],\notag
\end{equation}
where $l_{ij}$ is the element of the Jacobian of average system (\ref{eq:ave1})-(\ref{eq:ave5}) evaluated at equilibrium (\ref{eq:eq3}). We do not give the explicit form of $l_{ij}$ since it is quite lengthy and easily obtained. Both ${J^{eq3}}$ and ${J^{eq4}}$ has the same characteristic equation as the block diagonal matrix
\begin{equation}
\frac{1}{\omega }\cdot\rm{diag} \left\{
\left[ \begin{matrix}
{l_{22}}&{l_{23}}\\
{l_{32}}&{l_{33}}
\end{matrix}
\right],\;
\left[ \begin{matrix}
{l_{11}}&{l_{14}}&{l_{15}}\\
{l_{41}}&{l_{44}}&{l_{45}}\\
{l_{51}}&{l_{54}}&{-h}
\end{matrix}
\right]
\right\}. \notag
\end{equation}
The characteristic equation is
\begin{align}
0 &= \left[ {{s^2} - ({l_{22}} + {l_{33}})s + {l_{22}}{l_{33}} - {l_{23}}{l_{32}}} \right]\notag\\
&\quad\times\left( {{s^3} + {k_1}{s^2} + {k_2}s + {k_3}} \right), \label{eq:characteristic2}
\end{align}
where
\begin{align}
{k_1} &= h - {l_{11}} - {l_{44}},\notag\\
{k_2} &= {l_{11}}{l_{44}} - {l_{11}}h - {l_{44}}h - {l_{45}}{l_{54}} - {l_{14}}{l_{41}} - {l_{15}}{l_{51}},\notag\\
{k_3} &= {l_{11}}{l_{44}}h - {l_{14}}{l_{41}}h + {l_{11}}{l_{45}}{l_{54}} + {l_{15}}{l_{44}}{l_{51}}\notag\\
&\quad - {l_{14}}{l_{45}}{l_{51}} - {l_{15}}{l_{41}}{l_{54}}.\notag
\end{align}
According to the Routh-Hurwitz test \cite{routh1877treatise}, to guarantee that all the roots of characteristic equation (\ref{eq:characteristic2}) have negative real parts, we require the following conditions
\begin{align}
{l_{22}} + {l_{33}} < 0,\label{eq:cc21}\\
{l_{22}}{l_{33}} - {l_{23}}{l_{32}} > 0,\\
{k_1} > 0,\\
{k_2} > 0,\\
{k_3} > 0,\\
{k_1}{k_2} - {k_3} > 0. \label{eq:cc26}
\end{align}
The inequalities (\ref{eq:cc21}-\ref{eq:cc26}) are explicit stability conditions for Proposition 2, resulting in the following corollary:

\begin{corollary}
Consider the system (\ref{eq:sys1})-(\ref{eq:sys6}) with positive parameters $a$, $c_\alpha$, $c_\theta$, $b$, $h$ and $V_c$. Let these parameters are appropriately chosen to satisfy inequalities (\ref{eq:cc21})-(\ref{eq:cc26}), then for sufficiently large $\omega$, if the initial conditions ${r_c}(0),\theta (0),\alpha (0),e(0)$ are such that the following quantities are sufficiently small
\[\begin{array}{l}
\left| {\left| {{r_c}(0) - r^*} \right| - {{\rho _2}\sqrt {2{\gamma _3}} }} \right|,\left| {\alpha (0)} \right|,\left| {e(0) - {q_r}{R^2} - {e_2}} \right|, \\
\rm{and \quad either \quad} \left| {\theta (0)} - \arctan \frac{{{y_c} - {y^*}}}{{{x_c} - {x^*}}} - {{\mu _0}} \right| \\
\rm{or \quad} \left| {\theta (0)}  - \arctan \frac{{{y_c} - {y^*}}}{{{x_c} - {x^*}}} + {{\mu _0}} \right|,
\end{array}\]
then the trajectory of the vehicle center $r_c(t)$ exponentially converges to, and remains in the torus
\begin{align}
& \left| \alpha  \right|\le O(1/\omega ), \notag\\
{\rho _2}\sqrt {2{\gamma _3}} - O(1/\omega ) &\le \left| {{r_c} - {r^*}} \right| \le {\rho _2}\sqrt {2{\gamma _3}} + O(1/\omega ). \notag
\end{align}
\end{corollary}

\ifCLASSOPTIONcaptionsoff
  \newpage
\fi

\end{document}